\documentclass{article}
\usepackage{xcolor,colortbl}
\usepackage{color,soul}
\definecolor{Gray}{gray}{0.85}
\newcolumntype{C}{>{\centering\let\newline\\\arraybackslash\hspace{0pt}}m{0.1\textwidth}}
\newcolumntype{A}{>{\centering\let\newline\\\arraybackslash\hspace{0pt}}m{0.05\textwidth}}
\newcolumntype{V}{>{\centering\let\newline\\\arraybackslash\hspace{0pt}}m{0.4\textwidth}}
\newcolumntype{L}{>{\let\newline\\\arraybackslash\hspace{0pt}}m{0.85\textwidth}}
\newcolumntype{B}{>{\let\newline\\\arraybackslash\hspace{0pt}}m{0.82\textwidth}}
\newcolumntype{M}{>{\let\newline\\\arraybackslash\hspace{0pt}}m{0.1\textwidth}}
\usepackage{float}
\usepackage{comment}
\usepackage{placeins}
\usepackage{makecell}
\usepackage{array}
\usepackage{comment}
\usepackage[nolist]{acronym}

\usepackage{subfig}
\usepackage{amsmath}
\usepackage{enumitem}
\usepackage{longtable}
\usepackage{xcolor}
\usepackage{comment}

\usepackage{tcolorbox}
\usepackage{framed}

\usepackage{url}
\usepackage{enumitem}
\usepackage{placeins}
\usepackage{xcolor,colortbl}
\usepackage{color,soul}
\definecolor{Gray}{gray}{0.85}

\usepackage{float}
\usepackage{comment}
\usepackage{placeins}
\usepackage{makecell}
\usepackage{array}
\usepackage{comment}
\usepackage[nolist]{acronym}

\usepackage{amsmath}
\usepackage{enumitem}
\usepackage{longtable}
\usepackage{xcolor}

\usepackage{arxiv}

\usepackage[utf8]{inputenc} 
\usepackage[T1]{fontenc}    
\usepackage{hyperref}       
\usepackage{url}            
\usepackage{booktabs}       
\usepackage{amsfonts}       
\usepackage{nicefrac}       
\usepackage{microtype}      
\usepackage{lipsum}		    
\usepackage{graphicx}
\usepackage{natbib}
\usepackage{doi}

\usepackage{enumitem}
\usepackage[nolist]{acronym}
\usepackage{todonotes}
\usepackage{pdfpages}

\usepackage{color, soul}
\usepackage{makecell}
\usepackage{graphicx}
\usepackage[table]{xcolor}
\usepackage{xcolor}
\usepackage{soul,color}

\usepackage{longtable}
\usepackage{xcolor}
\usepackage{comment}

\usepackage{xcolor,colortbl}
\definecolor{Gray}{gray}{0.85}
\newcolumntype{C}{>{\centering\let\newline\\\arraybackslash\hspace{0pt}}m{0.2\textwidth}}
\newcolumntype{A}{>{\centering\let\newline\\\arraybackslash\hspace{0pt}}m{0.05\textwidth}}
\newcolumntype{V}{>{\centering\let\newline\\\arraybackslash\hspace{0pt}}m{0.4\textwidth}}
\newcolumntype{L}{>{\let\newline\\\arraybackslash\hspace{0pt}}m{0.75\textwidth}}
\newcolumntype{B}{>{\let\newline\\\arraybackslash\hspace{0pt}}m{0.82\textwidth}}
\newcolumntype{M}{>{\let\newline\\\arraybackslash\hspace{0pt}}m{0.1\textwidth}}

\newcolumntype{A}[1]{>{\raggedright\let\newline\\\arraybackslash\hspace{0pt}}m{#1}}

\newcolumntype{B}[1]{>{\centering\let\newline\\\arraybackslash\hspace{0pt}}m{#1}}

\renewcommand\hl[1]{#1}

\newcommand{\trtitle}{Disambiguating Anthropomorphism and Anthropomimesis in Human-Robot Interaction}
\title{\trtitle}

\author{
    Minja Axelsson\textsuperscript{\rm 1},
    Henry Shevlin\textsuperscript{\rm 1}\\
    \textsuperscript{\rm 1}University of Cambridge, UK  \\ 
 	\texttt{\{mwa29, hfs35\}@cam.ac.uk}
}

\renewcommand{\shorttitle}{Disambiguating Anthropomorphism and Anthropomimesis in Human-Robot Interaction}

\hypersetup{
pdftitle={\trtitle},
pdfsubject={},
pdfauthor={Minja~Axelsson, Henry~Shevlin},
pdfkeywords={Human-Robot Interaction, Social Robotics, Anthropomorphism, Anthropomimesis, Robot Design, User Perceptions, Human-like Robots},
}

\renewcommand\hl[1]{#1} 

\usepackage{bibentry}

\begin{document}

\maketitle

\begin{abstract}
In this preliminary work, we offer an initial disambiguation of the theoretical concepts \textit{anthropo\textbf{morphism}} and \textit{anthropo\textbf{mimesis}} in Human-Robot Interaction (HRI) and social robotics. We define anthropomorphism as users \textit{perceiving} human-like qualities in robots, and anthropomimesis as robot developers \textit{designing} human-like features into robots. This contribution aims to provide a clarification and exploration of these concepts for future HRI scholarship, particularly regarding the party responsible for human-like qualities---robot perceiver for anthropomorphism, and robot designer for anthropomimesis. We provide this contribution so that researchers can build on these disambiguated theoretical concepts for future robot design and evaluation. 

\vspace{0.3mm}

\end{abstract}

%

\section{Introduction}
\label{sec:introduction}

In this work, we set out to disambiguate and explore the theoretical concepts anthropo\textit{morphism} and anthropo\textit{mimesis} in the field of Human-Robot Interaction (HRI) and social robotics, in order to contribute to greater conceptual clarity in HRI scholarship. We argue that the key difference in these terms is regarding the responsible party for human-like features: in anthropomorphism, human-like qualities are perceived by the user, and in anthropomimesis, they are designed by the developer. \hl{The currently unclear distinction can result in the problem of unclarity of whether robot designers' choices, robot users' perceptions, or both, are contributing to the `human-likeness' of robots. As a consequence, it can be unclear how a robot's degree of `human-likeness' can be increased or decreased, and whether this should be done by changing design choices or addressing user perceptions. This in turn can lead to unclarity on how policy makers should respond to potential problems that result from anthropomorhipsm or -mimesis. Clarifying this distinction can shed light on how robots should be designed, how robot users should be informed, and how policy regarding robots should be shaped.} 
We discuss both anthropomimesis and anthropomorphism in relation to human-like robots, and set out to understand how these terms are currently used, and how we can clarify their definition in relation to the framing and interpretation of robots as human-like. 

Anthropomorphism is a widely used term in HRI \citep{damholdt2023scoping}, however, its definition is ambiguous. In a scoping review, \citet{damholdt2023scoping} found that across 57 studies ranging from 2000 to 2020, seven different definitions of anthropomorphism were used. While anthropomimesis as a term has not yet been popularised in HRI, we view it as being of interest for the HRI field, as it has been recently explored in philosophical literature regarding socially interactive and human-like Artificial Intelligence (AI) \citep{shevlinanthropomimetic}. Following the work by \citet{shevlinanthropomimetic}, we aim to make clear this same distinction when applied to HRI and social robotics, in order to clarify these theoretical concepts within the field. Our exploratory \textbf{research question} for this theoretical work is: \textbf{``How might we disambiguate and explore the theoretical concepts of anthropomorphism and anthropomimesis within the field of Human-Robot Interaction?''}

\section{Definitions of the Theoretical Concepts}

In this section, we set forth our working definitions of human-like robots, anthropomorphism, and anthropomimesis. A comparative analysis of the theoretical concepts anthropomorphism and -mimesis, as well as the responsible party, mechanism, and theoretical background related to each, is presented in Table \ref{table:comparative}. It is also important to note that our definitions do not foreclose each other. A robot that is perceived as anthropomorphic does not exclude the possibility that it was designed to be anthropomimetic, and a robot designed to be anthropomimetic does not prevent someone from perceiving it as anthropomorphic. In fact, we expect that anthropomimesis would have a positive effect on perceived anthropomorphism in most cases. However, in some cases, excessive anthropomimesis in a clearly non-human system may come across as ``fake'' or undesirable, leading to a negative effect on antorpomorphism. On the other hand, one can also anthropomorphise highly non-anthropomimetic systems.

\begin{table*}[]
\begin{tabular}{@{} p{1.5cm} p{1.5cm} p{4cm} p{8cm}   @{}}

\toprule
\textbf{Theoretical Concept}  & \textbf{Responsible Party} &  \textbf{Mechanism} &  \textbf{Theoretical Background}  \\ 
\toprule

Anthropo\textbf{-morphism}   
& Robot perceiver / user 
 
&  Users \textit{perceive} human-like qualities in robots
& Anthropomorhispm describes ``people attribut[ing] human characteristics to objects'' \citep{fink2012anthropomorphism}; ``Describes the human tendency to see human-like shapes in the
environment'' \citep{zlotowski2015anthropomorphism}; 
The tendency for humans to
attribute human qualities to non-human entities \citep{shevlinanthropomimetic}  

\\ 
\midrule

Anthropo\textbf{-mimesis}   
& Robot developer / designer 
 
& Robot developers \textit{design} features which mimic human-like qualities
& \textbf{Aesthetic anthropomimesis:} Mimetic robots are designed to resemble humans \citep{diamond2012anthropomimetic}; \textbf{Behavioural anthropomimesis:} The design and implementation of human-like
features in AI systems \citep{shevlinanthropomimetic}; \textbf{Substantive anthropomimesis:} The anthropomimetic principle describes a robot that ``imitates not just the human
form, but also the biological structures and functions that enable and constrain perception and action---and describes the design, construction, and initial performance of such a robot'' \citep{holland2006anthropomimetic};

\\

\bottomrule

\end{tabular}
\caption{Comparative analysis of anthropomorphism and -mimesis, identifying the responsible party, mechanism, and related theoretical background and definitions. } 
\label{table:comparative}
\end{table*}

\subsection{Human-Like Robots: ``Anthropos''}

``Anthropos''---the first concept in the words \textit{\textbf{anthropo}morphism} and \textit{\textbf{anthropo}mimesis}---
originates from the Greek word for ``human''. In this work, we are focused on ``human-like'' robots, that is, on the perception of human-like qualities in robots (i.e., anthropomorphism), and the design of human-like qualities into robots (i.e., anthropomimesis). By human-like qualities, we mean qualities of robots that resemble human qualities in terms of appearance, behaviour, or interaction. A human-like appearance constitutes a human-like body shape (with a head, a face, torso, arms and legs), while human-like behaviour and interaction involve socially interactive qualities, including engaging in human-like conversations and human-like behavioural signals and social norms. 
%
%
Our definitions of "anthropos"-like robots, in other words, human-like robots, encompass all human-like features of robots, including form-based qualities (i.e., physical and objectively observable qualities of a robot's appearance and embodiment), behaviours and interaction. 

It is also worth distinguishing our use of ``anthropos'' from other common and emerging terms in the HRI field. We focus here on \textbf{human}-likeness specifically---i.e., \textbf{anthropo}morphism and -mimesis. A related concept in HRI is zoomorphism and zoomorphic robots (e.g., \citep{fong2003survey, loffler2020uncanny, axelsson2021social}), which have ``animal-like'' qualities, or ``qualities [that] imitate living creatures'' \citep{fong2003survey}. Another related concept is sociomorphism and sociomorphic robots, i.e. ``the perception of actual non-human
social capacities'' according to \citep{seibt2020sociomorphing}. While similar arguments as those set forth in this paper might be made for the distinction between zoomorphism vs. zoomimesis and sociomorphism vs. sociomimesis, such arguments are out of scope for this paper. We encourage future research to explore the relation of this work's arguments to those potential explorations.

\subsection{Anthropomorphism: ``Morphe''}

The end of the word \textit{anthropo\textbf{morphism}} originates from the Greek word ``morphe'' for ``shape'' or ``form''. As briefly discussed in the Introduction, anthropomorphism as a term has a varied history throughout the HRI field, with \citet{damholdt2023scoping} identifying seven different definitions across 57 studies. Here, we detail the definitions of anthropomorphism they have identified, in order to relate them to the definition we put forth in this work. \citet{damholdt2023scoping} state that the most common definition is no definition given, and the second most common is definition through \citet{epley2007seeing}:``Imbuing the imagined or real behavior of nonhuman
agents with humanlike characteristics, motivations,
intentions, and emotions is the essence of anthropomorphism. These nonhuman agents may include anything
that acts with apparent independence, including nonhuman animals, natural forces, religious deities, and
mechanical or electronic devices (pp. 864-865).''. This is labelled the ``attribution-focused definition'', or definition 1 by \citet{damholdt2023scoping}. \citet{damholdt2023scoping} specify the next most common definitions: ``denoting the physical shape of the robot'' (``robotic feature -focused definition'', definition 2 \citep{damholdt2023scoping}), and the ``narrowed attribution-focused definition'', definition 3, which relies on the definition in the Godspeed Questionnaire Series by \citet{bartneck2009measurement}: ``Anthropomorphism
refers to the attribution of a human form, human characteristics, or human behavior to nonhuman things such as robots,
computers, and animals''.  \citet{damholdt2023scoping} provide further, idiosyncratic definitions identified from literature in their work, however while useful those are out of scope. 

Our definition of anthropomorphism relates most closely to definition 1 given by \citet{damholdt2023scoping} and originally set out by \citet{epley2007seeing}. However, our focus is on \textit{who} is doing the \textit{imbuing}: in our definition of anthropomorphism, imbuing is in the eye of the beholder, i.e., the user/perceiver of the robot. A definition more precisely in line with ours is that of \citet{fink2012anthropomorphism}, who state that ``people attribute human characteristics to objects'', in other words, users \textit{attribute to} and \textit{perceive} human-like qualities in robots, whether or not they can be objectively measured from the robot's design. We hold this to be distinct from anthropomimesis, in which human-like qualities are \textit{designed into} the robot by its developer/designer (whether intentionally or unintentionally---such a discussion of intentionality of the designer is out of scope for this work). This distinction is in line with the definitions set out by \citet{shevlinanthropomimetic}, where anthropomorphism is more straightforwardly defined as ``the tendency for humans to attribute human qualities to non-human entities'', and anthropomimesis as ``the design and implementation of humanlike
features in AI systems''. We will next discuss the definition of anthropomimesis in more detail.

\subsection{Anthropomimesis: ``Mimesis''}

The end of the word \textit{anthropo\textbf{mimesis}} originates from the Greek word ``mimesis'' for ``imitation''. \citet{shevlinanthropomimetic} makes a distinction between anthropomorphic and anthropomimetic non-embodied AI systems, in that anthropomimetic non-embodied AI systems are characterised by ``features of the system
itself'', rather than anthropomorphic systems which are characterised by ``user responses'' to that system (pp. 5). According to \citet{shevlinanthropomimetic}, the party responsible for the human-likeness in anthropomimesis is the system designer or developer, rather than the system perceiver or user in anthropomorphism (refer to Table \ref{table:comparative} for further analysis). While \citet{shevlinanthropomimetic} asserts that anthropomimetic features are consciously designed into non-embodied AI systems, we acknowledge that  anthropomimemtic human-like features may be consciously or unconsciously designed into a robot's form (i.e., appearance and embodiment), behaviour, or interactions. Anthropomimesis may be roughly described on three dimensions: aesthetic, behavioural, and substantive. Aesthetic anthropomimesis relates to physically observable qualities of the robot's embodiment, form and appearance (cf. \citep{phillips2018human}). Behavioural mimesis refers to robot behaviours which mimic human social and affective behaviours (cf. \citep{carpinella2017robotic}). Substantive anthropomorphism refers to mimicking the biological structures of the human body, including joints and muscle-like actuators \citep{diamond2012anthropomimetic}.

\citet{shevlinanthropomimetic} distinguishes between ``weak'' and ``robust'' anthropomimetic systems. They give the definition of a weak system as one where ```humanlikeness' is limited to surface-level features such as voice and
interface'', with the example of ELIZA. ELIZA was an early chatbot system from the 1960s, which used simple keyword identification and imitation to mimic human- and therapist-like conversations \citep{weizenbaum1966eliza}. A ``robust'' anthropomimetic system is described by \citet{shevlinanthropomimetic} as one which ``much more
closely resemble patterns of human cognition and behaviour'', with the example of ``contemporary LLM-based products like OpenAI’s
ChatGPT\footnote{https://chatgpt.com/} and Google’s Gemini\footnote{https://gemini.google.com/}''.\footnote{We acknowledge that these definitions are likely to change with time, and ELIZA may have been considered as robustly anthropomimetic at the time of its development and evaluation due to its relative sophistication in comparison to other systems at the time. However, we do not view this as a hindrance to our argument here as we are making our argument within a contemporary context, and as such consider discussion of this temporal context of changing interpretations to be out of scope for this work.} 
It is not entirely trivial to what extent such ``weak'' and ``robust'' anthropomimetic definitions of non-embodied AI can be applied to embodied AI systems, i.e., AI-enabled robots. In terms of robot behaviour and interactions, perhaps a ``robustly'' anthropomimetic robotic system would have an embedded LLM to drive its interactions and behaviours, while a ``weak'' anthropomimetic system would have a more simplistic conversational system such as ELIZA embedded to drive its interactions and behaviours. In terms of robot appearance and embodiment (i.e., form), the Uncanny Valley \citep{mori2012uncanny}---defined as ``the proposed relation between the
human likeness of an entity, and the perceiver’s affinity for it''---may be a useful concept for illustrating the distinction between anthropomimesis and anthropomorphism. Using the Uncanny Valley as a tool, ``robustly'' anthropomimetic robot systems could be those yet speculative robots which overcome the Uncanny Valley and resemble ``healthy persons'' and are practically indistinguishable from them, while ``weak anthropomimetic systems'' would lie prior to the valley in the realm of contemporary humanoid robots which are easily distinguished by the layperson from ``healthy persons'' by appearance (please refer to \citet{mori2012uncanny} for further definitions, discussion and figures on the Uncanny Valley).









\section{Discussion, Limitations and Implications}

In this section, we further discuss the disambiguation of the theoretical concepts of anthropomorphism and -mimesis, the limitations of our discussion, and its theoretical and practical implications.

\subsection{Discussion and Limitations}

While we have here set forth an initial disambiguation of the terms anthropomorphism and anthropomimesis on a theoretical level, we acknowledge that our definitions have limitations. \hl{This work encompasses an initial disambiguation of the concepts anthropomorphism and anthropomimesis, and in future work we aim to extend this work into a more detailed taxonomy of anthropomorphism and athropomimesis. Below, we detail more specific limitations of the present work, and future research to address those limitations.}

One of these limitations is the challenge of \textbf{measuring} anthropomorphism vs. anthropomimesis. Contemporary measures of anthropomorphism and anthropomimesis (e.g. the Godspeed questionnaire \citep{bartneck2009measurement} and the Anthropomorphic Robot or ABOT database\footnote{https://www.abotdatabase.info/}'s human-likeness measure) do not hold clearly distinct the distinction of these definitions. To demonstrate how these distinctions could be discretely measurable in the future, we argue that the measurability of such distinctions is easier to imagine when we take the example of robot form, or the physically observable and as such objectively measureable features of robot appearance. In terms of robot form, while the ABOT database focuses on ``physical human-likeness'' \citep{phillips2018human} and specifies objectively measureable human-like physical features on a yes/no categorical variable, e.g. ``Eyelashes: threadlike filaments around the eyelid'' and ``Nose: A projected feature of the face above the mouth'', they still describe this measure as follows: ``Robot Human-Likeness Predictor helps you forecast the extent to which your robot will be \textit{perceived} as human-like compared to other real-world robot''. Here, anthropomimetic features (eyelashes, nose) are used to predict the level of perception of human-likeness (i.e., in our definition, anthropomorphism). This ambiguous use of terminology limits the usefulness of the disambiguation of these theoretical concepts, as their distinct measurement may not be currently possible with state-of-the-art measures. \textbf{Future research} should set out to disambiguate the measurement of these distinct concepts, in order to further theoretical advancement and applicability of these concepts to future social robot and HRI design. \textbf{Future research} should also examine how the more challenging (i.e., not necessarily physically observable) qualities such as robot behaviour and personality can be measured in terms of perceived anthropomorphism and objectively observable anthropomimesis.

Another limitation is further exploring the \textbf{applicability} of this conceptual disambiguation to HRI research. An objection to this disambiguation may be that as anthropomorphism may be fundamentally predicated on anthropomimesis (i.e., perceptions of anthropomorphism are causally linked to anthropomimetic design choices), the usefulness of distinguishing anthropomorphism and anthropomimesis as concepts is limited for HRI and social robotics designers. However, we argue that understand the process through which a robot is perceived as anthropomorphism is useful, so that anthropomimetic cues can be chosen by an informed designer, for the desired effect. We also argue that this disambiguation can hold value due to the usefulness of assigning responsibility for perceived anthropomorphism. 
%
Responsibility for anthropomorphism might be more constructively assigned to users to the extent that it arose from potentially idiosyncratic user-level perceptions, whereas it would fall squarely to robot designers in cases where it followed directly from aggressively anthropomimetic design choices.
However, there may be cases where responsibility arises at the develop level even where anthropomimesis is limited and anthropomorphism is driven by user-side factors, such as vulnerability in the case of children or the cognitively impaired. Designers should consider such user-side factors, as well as potential cultural differences in what signals human-likeness, when designing anthropomimetic cues. Despite these challenges, identification of a responsible party for perceived anthropomorphism may be useful in the case of ``potential risks and harms arising from such responses'' \citep{shevlinanthropomimetic}. Understanding how a robot came to be perceived as anthropomorphic by a user, and whether and how this was elicited by anthropomimetic design choices by robot developers, is important for accountability and potentially even legal responsibility when interaction with human-like robots results in harm. Another argument for how this disambiguation holds value  is due to potential \textit{individual differences} in perceptions of anthropomorphism. The extent to which people perceive \textit{human-like qualities} in social robots may depend on their personal experience, demographics and preferences, while the same should not hold true for objectively measureable anthropomimetic features (e.g., having or not having a nose or eyelashes). For instance, \citet{perugia2022shape} found that gender of the perceiver impacts how ``feminine'' or ``masculine'' human-like robots were perceived as. We hypothesise that a similar effect might be found for individual differences in perceptions of overall anthropomorphism (rather than only gendered features), although confirmation of this would need \textbf{further empirical research}.

\subsection{Theoretical Implications}

This work set out to disambiguate the theoretical concepts of anthropomorphism and anthropomimesis in Human-Robot Interaction. To our knowledge, it is the first work attempting to do so. We aim for this work to contribute toward more robust theoretical definitions and discussions regarding anthropomorphism and -mimesis in HRI and social robotics. We view that robot designers and users both being aware of which party is responsible for human-likeness in a given robot design or application is important for creating useful and meaningful human-likeness future robot designs, and for the distribution of accountability in situations where human-like robots cause harm. We hope that our theoretical contribution can aid future examination of these important factors. 

\subsection{Practical Implications}

On a practical level, this work has implications for the design of HRI and social robots. HRI designers may wish to consider within their process when and how they are designing human-like features into their robots (i.e., anthropomimesis), and when and how any perceived human-like features are a result of user perceptions rather than design choices (i.e., anthropomorphism). Of course, both things may be true at once---for instance, a human-like robot design may elicit perceptions of human-likeness. Nevertheless, being aware of the difference in responsible parties may help robot designers become more aware of their human-like and non-human-like design choices, and robot users to become more aware of how their perceptions of a robot's human-likeness or non-human-likeness is shaped.

\section{Summary and Conclusion}

This work set out to disambiguate the theoretical concepts of \textit{anthropo\textbf{morphism}} and \textit{anthropo\textbf{mimesis}} in the field of Human-Robot Interaction (HRI). First, we discussed the definitions of and differences between the theoretical concepts of human-like robots, anthropomorphism, and anthropomimesis. Then, we discussed the limitations and implications of this theoretical disambiguation. This work aimed to contribute toward a more robust theoretical understanding of anthropomorphism (i.e., the perception of human-like qualities in robots) and anthropomimesis (i.e., the design of human-like qualities into robots). We highlighted the responsible parties for each concept:robot perceiver in the case of anthropomorphism, and robot designer in the case of anthropomimesis. We offer this contribution in order to contribute toward greater conceptual clarity, for the future design of HRI and social robots.

\section*{Acknowledgements}

M. A. is funded by the Emil Aaltonen Foundation. 
H. S. is funded by the Leverhulme Centre for the Future of Intelligence, Leverhulme Trust, under Grant RC-2015-067.

\section*{Open Access}
For open access purposes, the authors have applied a Creative Commons Attribution (\href{https://creativecommons.org/licenses/by/4.0/deed.en}{CC BY 4.0}) licence to any Author Accepted Manuscript version arising.




\clearpage

\bibliographystyle{plainnat}
\bibliography{main}

\end{document}